# Deep Learning Perspective of Scene Understanding in Autonomous Robots

Afia Maham
*Dept. of Computer Science*
*National Textile University*
Faisalabad, Pakistan
afiamaham08@gmail.com

Dur E Nayab Tashfa
durenayabtashfa@gmail.com

*Abstract*—This paper provides a review of deep learning applications in scene understanding in autonomous robots, including innovations in object detection, semantic and instance segmentation, depth estimation, 3D reconstruction, and visual SLAM. It emphasizes how these techniques address limitations of traditional geometric models, improve depth perception in real time despite occlusions and textureless surfaces, and enhance semantic reasoning to understand the environment better. When these perception modules are integrated into dynamic and unstructured environments, they become more effective in decision-making, navigation and interaction. Lastly, the review outlines the existing problems and research directions to advance learning-based scene understanding of autonomous robots.

## I. INTRODUCTION

Recent advancements in science and technology broadened the scope of autonomous robots in various fields health, logistics and manufacturing [1]. The tasks done by robots require advanced perceptual and cognitive abilities to interact with the dynamic and unstructured environment efficiently. So, the core part is understanding the three-dimensional (3D) environment, to perceive, interpret, and interact with it as humans do. This enables robots to make better decisions and interact meaningfully with humans and the environment.

Traditional computer vision approaches struggled with robustness, scalability and environmental adaptability due to their dependency on handcrafted features and geometric modeling. The transition into data-driven learning has modified how robots perceive their environments. With further development, deep learning now allows the autonomous robots to get rich hierarchical representations directly from raw sensory data. The improvement in feature extraction, contextual analysis, and multi-modal fusion by deep architectures, particularly Convolutional Neural Networks (CNNs), Recurrent Neural Networks (RNNs), and Transformers strengthens the robotic perception [2]. As a result, scene understanding has evolved from basic object recognition to thoroughly understanding environment. It forms the pillar of robust, real-time perception systems crucial for fully autonomous operations [3].

Many conventional techniques are bound to two-dimensional (2D) image domains, but the real world requires 3D perception which is still a cutting-edge research area. Autonomous robots need to interpret motion, spatial relationships, and environmental semantics instead of just relying on objects detection and recognition [1]. Moreover, most current robots such as service or cleaning robots are not yet fully autonomous, they perform only predefined or repetitive tasks. Manipulative and cooperative type complex tasks require deeper understanding of scenes that integrate geometric, semantic, and temporal information. Navigation in cluttered spaces, reason about dynamic objects, and adaptation to real world setting facilitated by this capability.

This review explores the basic principles, architecture, and empirical performance of the modern deep learning methodologies involved in scene understanding in autonomous robots. Furthermore, it carefully analyzes current limitations and challenges related to computational efficiency, data dependency, and ethical consideration. Also, the review discusses emerging directions such as AI-driven sensor fusion, quantum-assisted SLAM, and neural scene representations (e.g., NeRFs) that hold a potential to enhance the accuracy, adaptability, and reliability of robotic perception. Eventually, this paper gives understanding of the scenes in the light of deep learning, discussing the achieved progress and the persistent research gaps that continue to pave the way towards robotic systems that are truly autonomous.

## II. DEEP LEARNING FUNDAMENTALS FOR SCENE UNDERSTANDING

### A. Convolutional Neural Networks (CNNs)

Convolutional Neural Networks is a foundation of deep learning architecture for processing visual data. It can automatically capture hierarchical data patterns directly from raw pixel inputs through consecutive convolutions and pooling layers. Complex patterns, ranging from simple edges and textures to elaborate object parts and full semantic entities are identified by CNNs [4]. Their architectural design enables efficient parameter sharing and sparse connectivity which forms the basis of their success. It makes them expert at capturing spatial hierarchies in images and videos [5], [6].

CNNs can be used as a fundamental in many perception modules in navigation systems because of its strong command on feature extraction. It make the CNN-based models precious for tasks where understanding fine-grained spatial structure is important for avoiding obstacles and secure path planning. CNNs are integrated with various sensor modalities, such as RGB and LiDAR to enhance their capacity for accurate depth estimation and object recognition. Both are crucial for robust environmental perception in robotics [7].

Specifically, CNNs have enabled significant advancements in object detection, achieving high speeds and accuracy with models like Fast R-CNN, Faster R-CNN, SSD, and YOLO and in semantic segmentation. Largely driven by Fully Convolutional Networks and pyramid-based architectures that combine local and global image information [8].

### B. Recurrent Neural Networks (RNNs)

Recurrent Neural Networks are a distinct class of neural networks that possess the ability to process sequential data by maintaining an internal memory. It enables them to leverage information from previous inputs when maintaining the current ones [9]. This unique architectural feature involves connections that loop backward. They make RNNs ideal for tasks such as natural language processing, time series prediction, and video analysis by enabling them to capture temporal dynamics and dependencies [10] [11]. RNN has a wide use in NLP and time series analysis, but they are particularly valuable because of their ability to model temporal dependencies in perception tasks of robots.

RNNs are specifically beneficial for processing sequential sensor data, such as video streams or LiDAR scans over time, for estimation of future outcomes or identification of dynamic objects [12]. This capability is attained by intra-layer connections, establishing a feedback loop system inside the hidden layers [13]. RNNs leverage a hidden state, containing information from one step to the next, which enables them to model temporal dependencies. This behavior is unlike feedforward neural networks, which process inputs independently [14]. RNNs can determine context and make predictions based on series of observations through this memory mechanism. It is crucial for tasks like motion forecasting and behavioral prediction of other agents in a robotic environment [15].

Furthermore, variants like Long Short-Term Memory networks tackle the vanishing gradient problem inherent in standard RNNs. It enables them to capture longer-term dependencies which are crucial for understanding complex, extended sequences of autonomous system interactions and environmental changes [15].

### C. Generative Adversarial Networks (GANs)

Generative Adversarial Networks are a powerful class of deep learning models that learn the underlying distribution of training data. It enables them to generate new, synthetic samples which are indistinguishable from real data in most cases. This framework includes two neural networks, a generator that generates artificial data and a discriminator that differentiates between real and generated data samples. Both networks are simultaneously improving through an iterative training, so that they could reach a Nash equilibrium, where the generator produces highly realistic outputs [16]. This competitive learning framework allows GANs to generate data having strong statistical resemblance to the original dataset. It provides great facility to applications like data augmentation and the creation of realistic environmental simulations for training autonomous systems [9].

In robotic vision, GANs perform key roles in tasks such as super-resolution, image-to-image translation, and the generation of diverse training datasets. The main purpose is to improve the robustness of perception models in varied operational conditions [17]. By modeling normative scene distributions and identifying deviations, GANs provide a significant contribution in anomaly detection and uncertainty estimation [15].

Furthermore, GANs extend the functionality of other deep learning modules in scene understanding as they generate plausible representation for occluded, incomplete regions of environment. GANs are beneficial in scene understanding as they can provide a richer understanding of complex environments. The generation of unseen scenarios and filling in missing information are useful in occluded or partially observed scenes.

### D. Transformer Networks

To overcome the limitations of RNN in capturing long-range dependencies and parallelizing computations, transformer networks introduce the mechanism of self-attention to compute the importance of different parts of the input sequence [18]. Transformers, which were originally developed for natural language processing, have already made important breakthroughs in many fields of deep learning, including their increasing use in scene understanding as they adapted to robotic perception and control tasks. [19].

This is a non-recurring, non-convolutional architecture that uses multi-head attention to process entire input sequences at once. To enable a more global view of the contexts and relations in complex sensory data. They can be used in other robotic tasks, like perception, autonomous system planning, and control, with new versions of Transformers, including Vision Transformers and Detection Transformers [19] [20].

These models have been applied successfully to different tasks like object detection, semantic segmentation and even depth estimation with better results in responding to complex scenes by truly capturing long-range spatial relationships [21]. Transformer models demonstrated potential in improving accuracy of transparent object depth perception as a challenge that has a critical importance in robotic manipulation and most of the traditional 3D sensors have not been able to tackle effectively [22]. Their capacity to fuse information in various modalities and situational awareness makes them very appropriate in incorporating the multimodal sensor data into the autonomous driving systems [23].

III. OBJECT DETECTION AND SEMANTIC SEGMENTATION

*A. Object Detection Architectures*

The two broad categories of object detection models are the two-stage and the one-stage detectors, with the first group focusing on high accuracy by first giving regions of interest and then identifying and refining them, whereas the second group focuses on speed by completing all processes in a single run. Beyond traditional CNN-based detectors, transformer-based are used for capturing global context and long-range dependencies efficiently. Transformer- based architectures have also emerged as a recent development and use self-attention mechanisms to obtain state-of-the-art object detection performance through efficient capture of the global context and long-range dependencies [24] [25] [26] .

In particular, models as DETR and Deformable-DETR have demonstrated promising results, as these models rely on set-based loss functions for bipartite matching as well as enhance the convergence rate by using higher-spatial-resolution feature representations [27]. Subsequent work, including the TransVOD series merge object queries across frames by including temporal query encoders and deformable transformer decoders results in getting important temporal relationships to video object detection [27].

Transformer-based methods, including frameworks that represent 3D objects as queries have become prominent models of 3D object detection, especially in the case of data integration from many sensors [28] [29]. DETR3D and ORA3D models inspired by DETR, represent 3D objects as queries and utilize Transformer cross-attention capabilities, while PETR enhancing this result by constructing 3D position-sensitive representation [30].

To overcome limitations inherent in single-sensor systems and to enhance environmental comprehension, advance transformer architectures enable robust multimodal fusion which allows the integration of camera, LiDAR, and radar data [31]. Advanced transformer models include multimodal sensor fusion, integrating camera, LiDAR and radar data to improve robustness in complex settings.

*B. Semantic Segmentation Techniques*

In processing scenes, one of the most crucial tasks for autonomous robots is semantic segmentation, which assign each pixel of an image to a set of predefined categories to provide a dense, pixel-level knowledge of the surroundings. This contextual information is essential for complex robotic tasks like path planning, obstacle avoidance, and human-robot interaction which allows robots to understand scenes with accuracy [32].

Initially semantic segmentation relayed on region-based methods and graphical models, which are traditional computer vision techniques, but they faced challenges in feature extraction and contextual understanding of diverse environments. A fundamental change was brought about by the invention of deep learning, particularly Fully Convolutional Networks. They allow end-to-end learning of semantic segmentation directly from the pixels of an arbitrary image. These techniques are further refined by U-Net and DeepLab, which incorporate encoder-decoder architectures and atrous convolutions respectively. The main purpose is to capture multi-scale context and better boundary localization.

The transformer-based architecture, designed for natural language processing applications, has also shown encouraging performance on the task of semantic segmentation as it can capture the long-range relationships and global context in images [33]. These transformer-based architectures consist of approaches that encode dense image representations into range views, resulting in taking advantage of 2D segmentors in 3D operations [34].

*C. Instance Segmentation Approaches*

This classification that goes beyond just detecting the objects to the point where we can distinguish individual instances of objects within the same class, is essential in robotic interactions and manipulation processes. In contrast to semantic segmentation, where all pixels that represent a particular category are given the same label, instance segmentation does not offer uniform labels, but instead, each distinct object gets different label, which in turn enables the robots to locate and handle the individual objects in a cluttered environment [35]. Instance segmentation is required for more accurate manipulation and interaction tasks in robotic systems.

The ability is especially essential in the areas where distinguishing between several similar objects is required to carry out the required task accurately, such as grasping and manipulation. Instance segmentation algorithms often extend two-stage detectors with mask prediction branches, or they create one-stage models that predict instance masks directly [36]. As an example, Faster R-CNN is an extension of Mask R-CNN where the mask prediction branch is added; more recent methods, such as SOLO or CenterMask, convert the instance segmentation problem to a classification problem or integrate spatial attention-guided mask prediction branches respectively [36].

DeepLab series and SegFormer have also played an important role in enhancing the accuracy and efficiency of the segmentation. The former proposing atrous convolution and ASPP to do multi-scale target segmentation, and the latter use a hierarchical Transformer encoder to produce high-quality segmentation maps [37] [38]. Additional methods have worked on part-conscious segmentation to segment individual object instances with robustness particularly in industrial cases where objects are irregular and obscure [39].

### D. Applications in Autonomous Robotics

The integration of these advanced semantic and instance segmentation techniques is important for robotic navigation, intricate manipulation and interaction with humans safely [40]. Such technologies allow robots to discern and identify the nature of the surrounding objects, from pedestrians and vehicles to complex items in domestic or industrial settings, to make intelligent decisions and be responsive in dynamic environments. Combination of perception module with segmentation techniques allow robots to get better understanding of their environment which is essential for real-world applications.

Processes like autonomous driving and precision agriculture as well as medical robotics and surveillance, require the accurate environmental perception [41] [42]. Moreover, the combination of semantic and instance segmentation results with other sensor data, including LiDAR and radar, greatly increase the situational awareness of a robot and allow it to operate autonomously in unstructured and unpredictable settings more safely. For instance, in farming, instance segmentation allow robots to carefully delineate individual plants or fruits, which can then be used to perform tasks such as selective harvesting and precision pruning [41].

Such detailed segmentation is the basis of advanced robotic manipulation, which enables the precise grasping and handling of delicate objects by recognizing the individual component in a complex view [41] [43]. This allows robots to recognize objects and estimate poses even in varying illumination and occlusion environments [44]. Besides, it is observed that the performance of one-stage models such as YOLOv8 and two-stage ones such as Mask R-CNN vary in terms of performance trade-offs between inference speed and segmentation accuracy in different conditions, which is essential in real-time robotic applications [41] [43].

## IV. DEPTH ESTIMATION AND 3D RECONSTRUCTION

### A. Monocular Depth Estimation

The ill-posed task of monocular depth estimation aims at estimating the per pixel depth of a scene based on a solitary 2D image, offering vital spatial knowledge to an autonomous system [45]. Originally, techniques were based on geometric clues or handcraft features but with developments in deep learning, it is possible to learn depth maps end-to-end by using large datasets to boost accuracy and generalization. This capability is extremely crucial for autonomous navigation as robots must be able to calculate terrain shape and locate drivable areas so that the movements can be planned out efficiently [46].

The new deep learning methods extensively utilize convolutional neural networks to retrieve multi-scale features that are then combined to generate high-resolution depth maps and this has proven to be much better than the classical techniques because they learn complex correlations between image appearance and depth structure [47]. In addition, novel architecture, including the ones that embed self-supervised learning, reduce the requirement for ground-truth depth data, making them applicable to a wider range of scenarios in real-world. Hybrid models use neural networks with geometric models to solve tasks as robotic leaf manipulation, use methods like YOLOv8 to solve instance segmentation tasks and RAFT-Stereo to solve depth estimation tasks in 3D setting [48]. These methods have found beneficial in areas where active sensors are either infeasible or not available and provide a lightweight but robust real-time 3D perception system.

In addition, there has been increased use of attention mechanisms and transformer architectures in monocular depth estimation to generate global contextual information, results in achieving the consistency and accuracy of depth predictions in complex scenes [49].

### B. Stereo Vision and Multi-view Stereo

Stereo vision, which is a copy of biological vision, uses the method of triangulation of similar points in at least two images taken using different angles to determine depth, traditionally based on precise matching of features [50]. Traditional stereo methods often have trouble with textureless regions, occlusions, and require careful camera calibration to get precise and dense depth maps.

Modern deep learning approaches have tried to solve these issues by learning complex disparity-to-depth mappings directly from data [51] [52]. More recent developments of deep learning models of stereo vision include end-to-end learning of disparity maps, frequently with cost volume aggregation networks and regularization methods to address ambiguity, as well as to make them more robust to the demanding conditions of problematic situations [53]. Such approaches often combine neural networks to determine disparity or depth by building a cost volume and regularizing it with the help of 3D convolutions, such as in methods such as MVSNet [54].

Moreover, the combination of generative models, including Neural Radiance Fields, with multi-view stereo methods makes possible new synthesis of views and extremely detailed reconstruction of a scene in 3D by optimizing continuous volumetric scene functions [51]. It allows the development of extremely realistic virtual environments and in-depth 3D representations of sparse input images, which is important in applications such as augmented reality and simulation of autonomous vehicles [51].

Furthermore, the emergence of the so-called depth foundation models is a meaningful advancement that can provide generalized depth estimation in a wide variety of scenarios and sensor settings, just like in other fields, there are foundation models [55].

### C. LiDAR-based 3D Reconstruction

LiDAR-based 3D reconstruction is able to give highly accurate and dense point clouds of the environment, giving direct measurements of depth that are not sensitive to light changes unlike vision-based approaches [55]. Localization of proximate objects when primarily through LiDAR point clouds may be difficult, particularly in a complicated outside construction setting due to inherent sparsity of the point clouds [56].

Thus, more advanced methods of deep learning are being used to densify these point clouds and to generate these 3D structures, frequently by combining LiDAR data with data provided by a camera to exploit the advantages of each method [57]. Combination of LiDAR data with visual cues through multimodal learning may be used to solve the limitations of each modality alone by imputing lost depth data and semantically labelling the processed scenes. This fusion enables a more holistic and strong environmental perception, which is vital in activities like simultaneous localization and mapping in very dynamic backgrounds.

Moreover, alternative methods such as 3D Gaussian Splatting are becoming powerful to the conventional mesh or volumetric representation that provides high-quality three-dimensional reconstruction of sparse multi-view imagery with the ability of rendering in real time, which is useful in autonomous systems dynamic scene cognition [58]. These neural implicit representations give more effective and adaptable scene representation than the old approaches, overcoming such weaknesses of the old multi-view stereo methods as feature matching and low reconstruction [58].

### D. Neural Radiance Fields (NeRFs) for Scene Representation

Neural Radiance Fields have proven to be a novel innovation in scene representation. It enables the synthesization of novel views in a photo-realistic manner and constructs complex 3D scenes by widely training a continuous volumetric radiance field using 2D image input [59]. This method allows novel view generation with high-fidelity modeling and geometric consistency by mapping 3D coordinates and viewing directions to interpret them into color and transparency values. This was essentially based on optimizing this neural network to implicitly learn the geometry and appearance of the scene from the given collection of input images and camera poses [60]. Unlike traditional methods, the implicit representations are based on explicit geometries or voxel grids, which provide much more detail and continuity in the reconstructed scenes [61].

This technique has transformed 3D vision especially in tasks that involve generation of novel view and a complicated scene perception in robotics and augmented reality [62] [63]. Data inefficiency and training speed have become a common problem of NeRFs, as it requires intensive sampling of volumetric rendering, especially in sparse areas [64]. To reduce such challenges, later studies have investigated a variety of optimizations, such as explicit scene representation with sparse voxels, featured point clouds, and tensors, and with alternative network architectures, such as Multi-Level Hierarchies and infinitesimal networks, to make rendering faster and training more efficient [65].

## V. FUSING GEOMETRY AND SEMANTICS: VISUAL SLAM

### A. Traditional Visual SLAM Overview

The classic Visual SLAM systems usually estimate the camera pose and build a sparse or dense map of the environment simultaneously. It uses geometric feature matching and optimization schemes to ensure correct localization and the mapping in the unfamiliar surroundings. The methods used in these systems usually involve tools such as Kalman filters or bundle adjustment to improve pose estimates and map geometry, special focus on geometric consistency without any implicit semantic knowledge. Although this traditional methodology is robust in both pose estimation and sparse map generation domains, but it still struggle in semantic understanding, and consequently it is less useful when it comes to applications that need a higher-level understanding of the scene [66].

Conversely, Neural Radiance Fields provide a promising direction in SLAM by encoding the scene as the continuous volumetric function, rendering it as photorealistic as well as reconstructing the 3D scene in detail without necessarily extracting features [58]. Initial NeRF models had major computational drawbacks, such as a slow renderer, slow training, and could not be used in real-time in dynamic scenes [66] [67]. More recent work, including iNeRF and BARF integrate camera pose estimation directly in NeRF optimization process to deal with these limitations. It allows decoupling dependence on pre-computed poses and transitioning to a more full-fledged SLAM system [68]. These approaches typically use large Multi-Layer Perceptrons to represent the map and, therefore, the network inference can be extremely slow, or the training time can be very long [66].

B. *Semantic SLAM Approaches*

Semantic SLAM methods directly combine high-level scene perception with the localization and mapping process. It evolves from the purely geometric description of the scene to include object recognition, scene segmentation, and additional semantic information [69]. It enables robots to be aware of its surroundings which result in intelligent decision-making and interaction with complex environments [70]. Semantic SLAM leads to better mapping precision by identifying and classifying objects, especially in dynamic scenes where purely geometric methods got disrupted by moving elements [71].

This semantic enrichment results in better data association and closure of loops through the utilization of semantic similarity between observations. It enhances the global consistency and strength of the generated maps, especially when operating in challenging environments [72]. This can be used to generate more expressive and human-friendly map representations, and to have a more intuitive human-robot interaction where objects and locations may be described semantically instead of numerical coordinates [73]. Additionally, semantic information allows robots to exclude transient objects that make more stable and persistent maps which are essential in long-term autonomy [73]. Moreover, this combination of geometric and semantic information makes it feasible for robust localization in a setting where textures are low or where the setting consists of homogenous features since semantic landmarks can provide unique identifiers that cannot be determined through geometric devices [69].

C. *Deep Learning in Geometric SLAM Components*

Deep learning has radically changed the concept of geometric SLAM, no longer relying on conventional feature detection and matching techniques by using neural networks to accomplish activities of visual odometry, depth prediction, and resilient data association [66]. These deep learning techniques have shown impressive abilities to infer geometric properties directly using raw sensor data results in improving the quality and strength of pose estimation and map construction even under demanding conditions.

Deep learning systems can do real-time semantic segmentation and object detection to give at higher level information to SLAM systems to have a better perception of the environment around them [74] [75]. This semantic knowledge allows the robot to differentiate between static and dynamic features for more reliable and accurate localization and map management [71]. It has the potential to construct richer and semantically-aware maps that do not simply be a geometrically precise map but are also conceptually relevant to high-level robot tasks and long-term autonomy [73] [76]. Such synergy permits high-level autonomy, since robots can use semantic representations to make better decisions and interact with the environment in better way [73]. However, introducing semantic information into classical SLAM paradigms is a problem because of no common mathematical framework to integrate semantic and geometric estimations into one system [73].

VI. DYNAMIC ENVIRONMENTS

A. *Understanding Dynamic Objects*

The natural variability of dynamic objects, as well as their unpredictable pattern of motion, poses considerable problems to autonomous robots to determine the accurate perception and interaction of their environment. Conventional SLAM systems tend to view dynamic objects as outliers which are filtered by robust estimation algorithms. It can cause information loss and incorrect view of the surroundings [73]. As a result, the differentiation of static backgrounds and the dynamic foreground element becomes highly important for proper and robust localization and mapping in dynamic scenes. This distinction is not only essential for localization accuracy but also is essential for safer navigation and interaction in environments having autonomously moving objects.

Thus, precise recognition and monitoring of dynamical objects is essential to prevent collisions and enable efficient path planning in complicated and realistic environments [71]. However, the inherent problem of whether it was the robot or the environment that is in motion remains a strong obstacle in the process that is usually complicated by the fact that observability is lacking in most real-life situations [73]. To deal with this, more advanced methodologies usually use advanced motion models and data association algorithms to detect motion of robots from changes in environment and use multi sensor fusion to provide greater robustness [69]. Moreover, by incorporating the features of deep learning, the capability to locate and monitor dynamic objects, even when cluttered/partially occluded, has greatly enhanced, and it utilizes learned features and motion patterns to achieve it [71].

### B. Motion Prediction for Autonomous Navigation

Motion prediction of dynamically moving objects is an essential factor of autonomous navigation as it allows the robot to predict the future of the movements of other agents in the environment and execute collision-free paths [73]. This proactive capability converts reactive collision avoidance-based robot planning into predictive interaction, which is essential for smooth operation in very interactive and unpredictable settings. With knowledge of dynamic changes, robots can produce proactive plans and actions, thus increase coordination and enable behavior adaptation particularly in activities that require human interactions [77]. This is especially important in complicated situations where most commonly, traditional navigation pipelines are substituted by a reinforcement learning approach, yet self-supervised learning can also play an important role in several tasks, such as the ability to predict the future movement of dynamic obstacles [78]. This will enable the robots to acquire self-directed learning and adjust to different forms of environment and interaction and will improve their capability to move in complex real-world environments [79].

Also, object tracking, especially the multi-object tracking, is crucial to enable robots to prevent collisions by adjusting their navigation plans [80]. These methods such as DepTR-MOT are able to combine instance-level depth clues and can be used to filter trajectories in multi-object tracking which results in overcoming the weaknesses of 2D information in highly dense robotic scenes [81]. Furthermore, to understand dynamic environments, the integration of perception and prediction modules offers a more comprehensive approach. It moves beyond independent processing streams to jointly optimize these critical functions for autonomous systems [82].

### C. Robust Scene Understanding in Changing Conditions

Self-supervised learning methods are becoming a powerful paradigm to facilitate dynamic object detection and segmentation without large and annotated datasets. It results in providing scalable solutions to a wide variety and ever-changing operational settings [83]. Spatio-temporal occupancy grid map includes information regarding the future of dynamic scenes. This approach enables autonomous systems to predict them which allows risk assessment in real-time and proactive planning of pathways [79]. This would enable the robots to navigate around complicated environments by understanding the movement of dynamic obstacles and incorporate the predictions into their navigation frameworks [79].These predictive capabilities are essential for safe and efficient operation, especially in situations having multiple interacting agents, where object-centric methods may difficult to scale [79].

Transformer architectures including NavFormer have been built to help target-driven navigation in dynamic and unknown environments. They can do this by sequential data processing and self-supervised visual representation to reason about spatial layouts and support collision avoidance [84]. It may be used to do an end-to-end learning strategy, where navigation decisions are conditioned on both target images and history of robot paths, which improves adaptability in new dynamic conditions [84]. Moreover, multi-modal data (the combination of RGB images and LiDAR) integration is becoming highly important to improve perception capabilities and system robustness in complicated settings [85].

## VII. CHALLENGES AND FUTURE DIRECTIONS

### A. Real-time Performance and Computational Efficiency

The main challenge in autonomous robots during implementation of deep learning models for dynamic scene understanding is to provide real-time work on embedded devices with few computing units [86]. It requires the construction of highly optimized architectures and efficient inference methods that are capable of maintaining accuracy while performing within stringent power and processing limits. Various techniques are being explored including model quantization and pruning, and neural architecture search to simplify models without significantly compromising their performance, to enable trained models to be deployed on resource-constrained robotic systems [86].

### B. Robustness to Illumination and Environmental Variations

Autonomous robots usually face diverse and challenging environments such as illumination changes to severe weather conditions, all of which severely affect the perception system. Therefore, researchers are still focused on establishing strong algorithms which can be capable of maintaining a high performance over a large range of illumination levels, occlusions, and dynamic transitions. Adversarial training and domain adaptation are being investigated by novel approaches to generalize its models to unknown conditions results in avoiding the need to collect large amounts of real-world data on every possible condition. Moreover, novel generative architectures are also under consideration as a way of generating missing frames within video streams, which can become one of the potential remedies to lack of information and sensor failures in self-driving scenarios [87].

### C. Explainability and Interpretability of Models

To build trust and facilitate successful human-robot interaction, having the rationale of the decisions that a robot makes, especially in dynamic and safety-critical scenarios, is most important. This creates a need to develop learnable deep learning models which are able to give us an insight on how they make decisions and not just using black box learning methods but instead give them transparency and accountability. This kind of advancement plays a vital role in both debugging and in making certain that autonomous systems are applied ethically, particularly in highly unstructured situations where unexpected interactions are common. Thus, the creation of ways to view and describe the qualities that deep networks learns and their impact on the behavioral outputs is an expanding field. Moreover, ethical and legal concerns of an autonomous decision-making should depict the models not only robust but also complying with the existing norms and regulations, especially in the area of privacy and safety [88].

### D. Data Efficiency and Synthetic Data Generation

One of the major issues of deep learning in robotics is the sheer annotated data for training that is costly and may be laborious to obtain, particularly with heterogeneous and changing real-world situations [89]. This requires new data augmentation methods, self-supervised learning, and creation of high-fidelity synthetic data to alleviate the need to use labor-intensive manual labeling [90]. In particular, generative adversarial networks, and other probabilistic modelling methods, are being used to generate naturalistic synthetic datasets, which can help close the sim-to-real gap, results in faster model creation and deployment in autonomous systems [91] [92].

### E. Ethical Considerations and Safety Implications

As the autonomy of robotic systems is increasing, it brings profound ethical concern about responsibility in the event of accidents and the potential for algorithmic bias [93]. To develop trustworthy and ethically aligned robotic systems, it is important to ensure transparency in AI decisions. It promotes accountability, informed consent, and the debugging of ethical algorithms [94]. This involves coming up with clear regulations as well as effective certification measures to evaluate and ensure the safety, fairness and privacy of these opaque systems particularly since they actively interact with the real world [95] [96].

### VIII. Conclusion

Looking at the challenges and new directions in perception of autonomous robots, it is essential to consider the overall progress and remaining research gap in the fields. The paper has critically looked at the progress and challenges in scene understanding for autonomous robots, showing the paramount importance of deep learning in different perception tasks. From object detection and semantic segmentation to depth estimation, 3D reconstruction, and visual SLAM, deep learning has significantly enhanced the capabilities of robots to interpret and interact with complex environments. Even after so much progress, still significant research gaps remain in achieving real-time performance. It ensures robustness in various environments and generates explainable and ethically sound models for dynamic and safety-critical robots. Thus, future research should aim at novel mechanisms of achieving a balance between speed and capacity when making decisions in real time, especially complex VLA models, with the aim of producing the right actions at the right time in an ever-changing environment.